\documentclass[letterpaper, 10 pt, conference]{ieeeconf}  

\IEEEoverridecommandlockouts                              

\usepackage{graphics} 
\usepackage{amsmath} 
\usepackage{cite}
\usepackage{amssymb}  
\usepackage{multicol}
\usepackage{pgfplots}
\usepackage{subfigure}
\usepackage{float}
\usepackage{multirow}
\usepackage{url}
\usepackage[flushleft]{threeparttable}
\usepackage{algorithm}
\usepackage{algorithmicx}
\usepackage{algpseudocode}
\usepackage{tikz}
\usetikzlibrary{shapes,arrows}
\usepackage{mars}

\usepackage{verbatim}

\algdef{SE}[DOWHILE]{Do}{doWhile}{\algorithmicdo}[1]{\algorithmicwhile\ #1}%

\title{\LARGE \bf
Path Planning Tolerant to Degraded Locomotion Conditions
}

\author{Xiaoling Long$^{1}$ and S\"oren Schwertfeger$^{1}$
\thanks{$^{1}$Both authors are with School of Information Science Technology of ShanghaiTech University
        {\tt\small <longxl, soerensch>@shanghaitech.edu.cn}}%
}

\begin{document}

\marsPublishedIn{Accepted for:} 		

\marsVenue{IEEE INTERNATIONAL SYMPOSIUM ON SAFETY, SECURITY AND RESCUE ROBOTICS 2019}

\marsYear{2019}

\marsPlainAutors{Xiaoling Long and S\"oren Schwertfeger}


\marsMakeCitation{Path Planning Tolerant to Degraded Locomotion Conditions}{SSRR}

\marsDOI{ 10.1109/SSRR.2019.8848980}

\marsIEEE{}


\makeMARStitle

\tikzstyle{block} = [draw, fill=blue!20, rectangle, text width=7.5em,
    minimum height=6em, minimum width=4em, node distance=5.5cm]
\tikzstyle{obstacle} = [draw, fill=black, rectangle, text width=7em,
    minimum height=1em, minimum width=3em, node distance=5.5cm]
\tikzstyle{sum} = [draw, fill=blue!20, circle, node distance=1cm]
\tikzstyle{input} = [coordinate]
\tikzstyle{output} = [coordinate]
\tikzstyle{pinstyle} = [pin edge={to-,thin,black}]
\tikzstyle{line} = [draw, -latex']
\tikzstyle{cloud} = [draw, ellipse,fill=red!20, node distance=2.5cm,
    minimum height=2em]
\tikzstyle{decision} = [diamond, draw, fill=blue!20,
text width=5em, text badly centered, node distance=2cm, inner sep=0pt]
\tikzstyle{block1} = [rectangle, draw, fill=blue!20,
    text width=5em, text centered, rounded corners, minimum height=2em]

\maketitle
\thispagestyle{empty}
\pagestyle{empty}

\begin{abstract}

  Mobile robots, especially those driving outdoors and in unstructured terrain, sometimes suffer from failures and errors in locomotion, like unevenly pressurized or flat tires, loose axes or de-tracked tracks. Those are errors that go unnoticed by the odometry of the robot. Other factors that influence the locomotion performance of the robot, like the weight and distribution of the payload, the terrain over which the robot is driving or the battery charge could not be compensated for by the PID speed or position controller of the robot, because of the physical limits of the system. Traditional planning systems are oblivious to those problems and may thus plan unfeasible trajectories. Also, the path following modules oblivious to those problems will generate sub-optimal motion patterns, if they can get to the goal at all.

In this paper, we present an adaptive path planning algorithm that is tolerant to such degraded locomotion conditions. We do this by constantly observing the executed motions of the robot via simultaneously localization and mapping (SLAM). From the executed path and the given motion commands, we constantly on the fly collect and cluster motion primitives (MP), which are in turn used for planning. Therefore the robot can automatically detect and adapt to different locomotion conditions and reflect those in the planned paths.

\end{abstract}

\section{INTRODUCTION}
Robotic motion planning is an important topic in mobile robotics. How to reach the goal and how to avoid collisions are two key challenges. In current robot motion frameworks, a global planner tells the robot how to reach the goal. A local planner makes the decisions how to move to follow the planned path and avoid collisions at the same time.
\par
For the global planner, many graph search technologies could be applied to the search space~\cite{lavalle2006planning}, \textit{ie, A* and Dijkstra's algorithm in grid search space}. Both of them can find the optimal path if no time and memory limitations are given. Extensions \textit{eg, D*\cite{koenig2002d}, ARA*\cite{Likhachev2003ARAAA}}  can speed up the planning. The normal graph search only considers the robot position, but ignores the orientation and other states of the robot. Non-smooth and discontinuous paths cannot be followed by real physical robots. The elastic bands model is able to deform the non-smooth and discontinuous paths into short and smooth paths, which the robots are able to follow~\cite{quinlan1993elastic}. \cite{kurniawati2007path} argue that path deformation might fail in dynamic environments, so state-time space is used to deform the path to maintain the obstacle clearance and connectivity of the path. It proposes that trajectory deformation is more suitable for robots in dynamic environments. Kinodynamic motion planning also considers the smoothness and connectivity of the global path. It is using splines to optimize the path to yield the smooth and time-optimal trajectories.
The key challenge of differential drive robots is, that they do not have the ability to move sideways, so planning without taking the orientation of the robot into account can lead to non-feasible or sub-optimal paths.
\par
\begin{figure}
  \centering
  \includegraphics[width=0.4\textwidth]{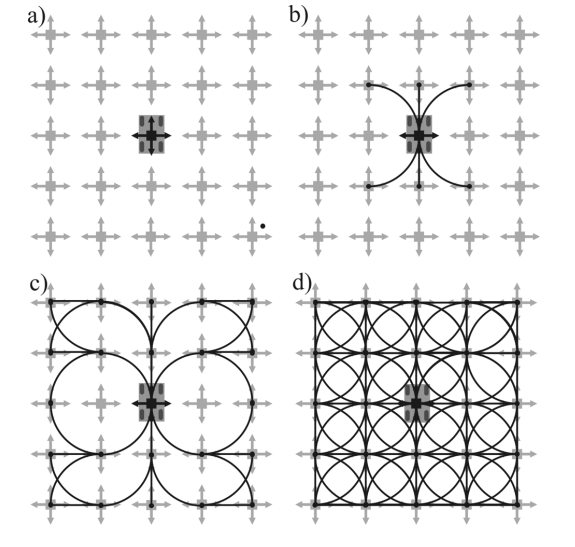}
  \caption{Constructing the lattice for the Reeds-Shepp Car.}
  \label{statelattice}
\end{figure}

\cite{pivtoraiko2005efficient} proposes an efficient path planning method in state lattices. Instead of a search in square lattices, it constructs a special search space which encodes feasible paths for differential drive robots. Any resulting path from this method satisfies the differential drive constraints. Figure~\ref{statelattice}  shows the state lattice used for the path planning~\cite{pivtoraiko2005efficient}. \cite{1545046} propose a method to generating minimal spanning control sets for state lattice.
\par
A global planner provides guidance for the navigation task, while a local planner decides the actions of the robot for following the global path. The dynamic window approach (DWA) performs a one-step simulation using velocity sampled
 from the set of achievable velocities and selects the highest-scoring result to execute\cite{fox1997dynamic}. Trajectory Rollout samples from the set of admissible velocities over the entire forward simulation period under the limitation of acceleration of robot\cite{gerkey2008planning}.

\par
In a search and rescue scenario, the mobility of the robot varies in different environments and also depends on the robot state (e.g. payload, battery charge, wheel/ track condition). Especially in the difficult terrain often present in search and rescue scenarios, wheel slip may occur quite often, such that the odometry is unreliable~\cite{ward2007model}. In this paper we are considering circumstances that may change the dynamic or kinematic of a robot motion. For example wheels may loos pressure or be punctured, an axis or chassis might bend, a payload might be placed or shifted resulting in a shifted center of gravity. Mechanical failures such as de-tracking or loose couplings between wheels and axis also effect the motion of the robot.
\par
Those changes in robot motion may be captured by the odometry (e.g. low battery charge resulting in less torque; shifted center of gravity) or the odometry may be oblivious of the real robot motion (e.g. diameter change of tires due to loss of pressure; de-tracking on one side). The proposed algorithm will be able to handle all those cases by making use of external localization in the form of Simultaneous Localization and Mapping (SLAM).

\par
In this paper, we adopt a planning method that is using motion primitives that are obtained from the actual robot motion. It produces smooth paths and automatically learns the robot's kinodynamic constraints. The motion primitives span the whole achievable work space of the robot. The highlight of this paper is, that the motion primitives based path planning is tolerant to degraded locomotion conditions, because our algorithm continuously samples the real robot motion, generates motion primitives and clusters them for planning.

The rest of the paper is organized as follows: Section~\ref{sec:mpbplaning} introduced our motion primitives based path planning. Section \ref{sec:experiments} presents two experiments about our algorithm and the last section concludes this paper.

\section{Motion Primitives Based Planning}
\label{sec:mpbplaning}

We treat path planning as a graph search problem. Grid maps are a commonly used 2D map representation. In normal path planning, a 4-connected or 8-connected grid map is used for graph search.
\begin{figure}[b]
  \centering
    \begin{tikzpicture}
      \centering
      \draw[step=1cm, gray, very thin] (-1.4, -1.4) grid (1.4,1.4);

      \draw[very thick, ->] (0,0) -- (1,0);
      \draw[very thick, ->] (0,0) -- (-1,0);
      \draw[very thick, ->] (0,0) -- (0,1);
      \draw[very thick, ->] (0,0) -- (0,-1);
    \end{tikzpicture}\quad
    \begin{tikzpicture}
      \centering
      \draw[step=1cm, gray, very thin] (-1.4, -1.4) grid (1.4,1.4);

      \draw[very thick, ->] (0,0) -- (1,0);
      \draw[very thick, ->] (0,0) -- (-1,0);
      \draw[very thick, ->] (0,0) -- (0,1);
      \draw[very thick, ->] (0,0) -- (0,-1);
      \draw[very thick, ->] (0,0) -- (1,1);
      \draw[very thick, ->] (0,0) -- (1,-1);
      \draw[very thick, ->] (0,0) -- (-1,1);
      \draw[very thick, ->] (0,0) -- (-1,-1);
    \end{tikzpicture}\\
    %
    %
  \caption{\textbf{Left}: 4-connected lattice grid map. \textbf{Right}: 8-connected lattice grid map.}
  \label{gridmap}
\end{figure}
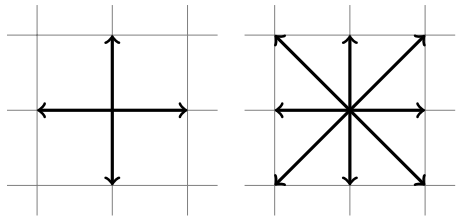

\begin{figure}[t]
  \centering
  \begin{tikzpicture}[auto]
    \draw[step=1cm, gray, very thin] (-1.4, -1.4) grid (1.4,1.4);

    \draw[very thick, ->] (0,0) to [bend left=45] (1,1);
  \end{tikzpicture} \quad
  \begin{tikzpicture}[auto]
    \draw[step=1cm, gray, very thin] (-1.4, -1.4) grid (1.4,1.4);

    \draw[very thick, -] (0,0) to [bend left=45] (0.5,0.5);
    \draw[very thick, ->] (0.5,0.5) to [bend right=45] (1,1);
  \end{tikzpicture}
  \caption{Smooth path to (1, 1). \textbf{Left}: start heading up and end heading right. \textbf{Right}: start heading up and end heading up.}
  \label{smooth}
\end{figure}
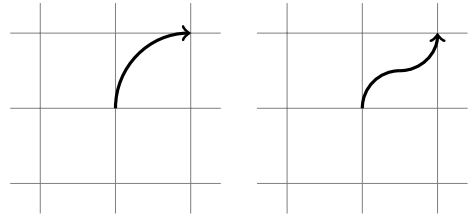

Both 4-connected and 8-connected lattice grid maps result in discontinous heading direction, which is not executable for mobile robots except omni-robots. For differential robots, a smooth and continous curve is required, as Figure \ref{smooth} shows, indicating that the start heading is up and end heading is right. If robot needs to end heading up, it might need take other action such as Figure \ref{smooth} right.

The path in Figure \ref{smooth} is a lattice-based motion primitive\cite{pivtoraiko2005efficient}. \cite{Pivtoraiko2012GeneratingSL} compares the control-sampling primitives and the state lattice motion primitives. The state lattice primitives has an advantage as 8-connected grid map. The vertexes locate on the original vertex positions and are highly aligned. Forward motion primitives can be obtained easily, and can be easily updated online. Reverse motion primitives (state lattice motion primitives) have aligned vertexes, this type of primitives obtain the corresponding robot action by using a Boundary Value Problem (BVP) solver\cite{Pivtoraiko2012GeneratingSL}, making it not easy to maintain and update.
\begin{figure}[tb]
  \centering
  \includegraphics[width=0.5\textwidth]{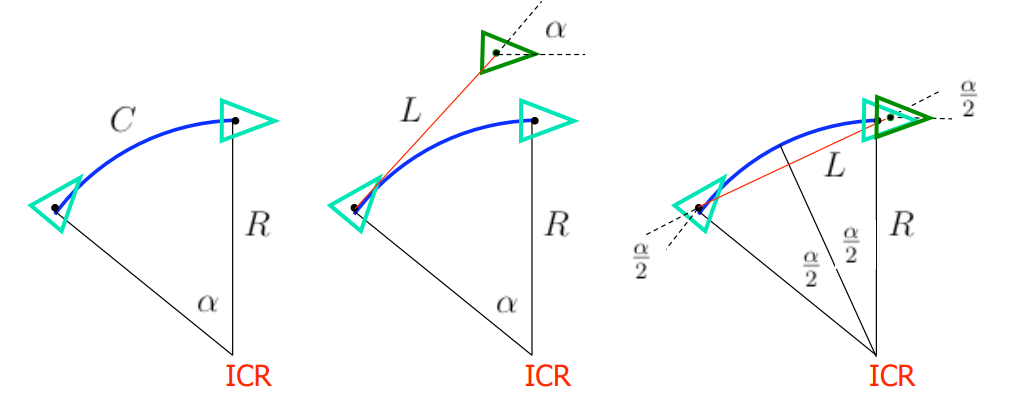}
  \caption{Calculating Forward Kinematics: \textbf{Left}: Exact odometry via integration;
  \textbf{Middle}: Euler approximation; \textbf{Right}: Second order Runge-Kutta appromaxition.}
  \label{odom}
\end{figure}

Figure \ref{odom} shows more information about the two kinds of approximation \cite{Odom}. The control-sampling motion primitives are collected by generating control signals and using forward kinematics to calculate the resulting robot motion and pose.
Generating control-based primitives by approximation has issues of  inaccuracy and it is difficult to generate complex primitives. Therefore, obtaining the motion primitives from the real motion will be more efficient and accurate.
Each control sample corresponds to one specific path of the robot in the environment. Generally, a approximate result of control-sampling motion primitives can not represent the real robot motion. It also varies in different workspaces.
\par

%
%

\begin{figure}[b]
  \begin{tikzpicture}
    \centering
    \node [block, name=step1] {Generate and update motion primitives};
    \node [block, right of=step1] (step2) {Path Planning in the  MP-based search space (A*)};
    \node [block, node distance=3cm, below of=step2] (step3) {Execute the plan and evaluate the behavior of execution};

    \draw [draw, ->] (step1) -- (step2);
    \draw [->]  (step2)  -- (step3);
    \draw [->]  (step3)  --  (step1);
  \end{tikzpicture}

  \caption{Diagram for Motion Primitive (MP) -based Planning}
  \label{flow}
\end{figure}
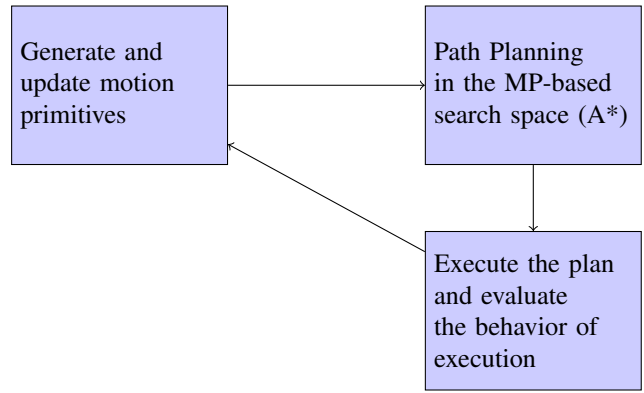

Figure~\ref{flow} shows the overview of our approach. In the first step we generate motion primitives. For that we record the control inputs and observe the resulting robot motions using SLAM for localization. From the collected samples we use a clustering algorithm to select a few representative Motion Primitives (MP). In the second step we use the MPs for planning. A* is utilized to quickly find a path to the goal. In the third step the planned path is executed by simply re-playing the control input of the selected motion primitives. Here we also use the localization from SLAM to evaluate the performance of the MPs (i.e.: Did we end up at the pose we expected?). The evaluation result will then influence which clusters of MPs are enabled in step one and thus available for the path planning in step two.

\subsection{Generate the Motion Primitives}

Motion Primitives (MPs) can be generated from any robot motion. The motion can come from tele-operation, another navigation stack or the navigation algorithm presented here.

As shown in Algorithm~\ref{alg:mpg}, in any case, we sample the control input as well as the resulting robot motion. The robot motion is obtained from the localization result of a SLAM algorithm. The control input can be in any form, for example raw motor speeds or twist messages, which encode a desired robot motion (translation and rotation speeds).

\begin{algorithm}[b]
  \caption{Generate Motion Primitives}
    \label{alg:mpg}
  \begin{algorithmic}[1] 
      \Require Control command $c(t)$, SLAM real-time position $p(t)$, Time interval $\Delta t$
      \Ensure Motion Primitives $M$
      \Function {MotionPrimitivesGenerator}{}
      \State $t_{0} =$ getTime()
      \State $P, C, T = \emptyset$
      \While {$c(t)$ not zero}
          \State $t= $ getTime()
          \If {$t - t_{0} > \Delta t$}
            \State  $M$.append([$C$, $T$: $P$])
            \State $P, C, T = \emptyset$
          \Else
            \State $C$.append($c(t)$)
            \State $T$.append($t - t_{0}$)
            \State $P$.append($p(t)$)
            \State $t_{0} = t$
          \EndIf
      \EndWhile
      \State \Return{$M$}
      \EndFunction
  \end{algorithmic}
\end{algorithm}

%

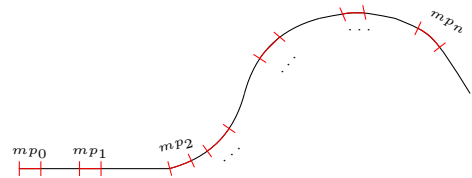
\begin{figure}[b]
  \centering
  \begin{tikzpicture}[auto]
    \draw (0,0) to  (2,0);
    \draw[|-|, red] (0,0) to (0.3,0);
    \node[] at (0.15, 0.2)
    {\tiny $mp_{0}$};
    \draw[|-|, red] (0.8,0) to (1.1,0);
    \node[] at (0.95, 0.2)
    {\tiny $mp_{1}$};
    \draw (2,0) to [bend right] (3,1);
    \draw[|-|, red] (2,0) to [in=-155, out=15] (2.3,0.11);
    \node[rotate=15] at (2.1, 0.3)
    {\tiny $mp_{2}$};
    \draw[|-|, red] (2.5,0.23) to [in=-125, out=36] (2.8,0.54);
    \node[rotate=40] at (2.85, 0.2)
    {\tiny $\cdots$};
    \draw (3,1) to [bend left] (4,2);
    \draw[|-|, red] (3.2,1.46) to [in=-130, out=57] (3.47,1.76);
    \node[rotate=50] at (3.6, 1.4)
    {\tiny $\cdots$};
    \draw (4,2) .. controls (4.5,2.1) .. (5,2)
      (5,2) .. controls (5.5,1.8) .. (6,1);
    \draw[|-|, red] (4.3,2.05) to [in=-168, out=23] (4.6,2.08);
    \node[rotate=0] at (4.55, 1.85)
    {\tiny $\cdots$};
    \draw[|-|, red] (5.3,1.87) to [in=128, out=-25] (5.6,1.61);
    \node[rotate=-25] at (5.7, 1.95)
    {\tiny $mp_{n}$};
  \end{tikzpicture}
  \caption{Example for generating motion primitives from robot motion. $mp_{i}$ indicate the generated $i$th motion primitive}
  \label{path}
\end{figure}

Figure~\ref{path} shows an example of how to generate lots of motion primitives from an executed robot motion. These motion primitives do not just include the result path, but also include the corresponding control input.
There will be lots of similar MPs, i.e. the $mp_{0}$ and the $mp_{1}$ are very similar - in path and control input. Therefore the generated motion primitives have lots of redundant information. We should not use all of those primitives from planning, as it would need huge computational resources without gaining any advantage. So we adopt a clustering algorithm to select a few MPs representative for all motions of the robot. We use affinity propagation~\cite{frey2007clustering} to provide the cluster center from the datasets, while K-means and mean shift iteratively compute the centeroids of the clusters.

\subsection{Path Planning in the Motion Primitives based Search Space}

\begin{figure}[tb]
  \includegraphics[width = 0.5\textwidth]{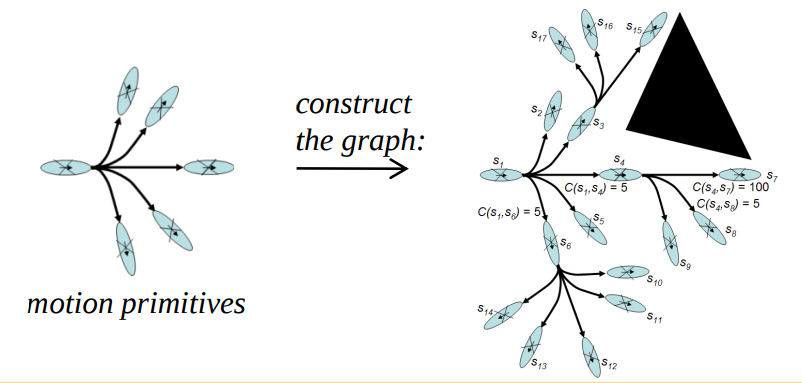}
  \caption{Path planning with motion primitives.}
  \label{ss}
\end{figure}

Given motion primitives, the problem is how to find the optimal  path from a start state to a target state. The optimality includes not only the minimal length of the path but also the smoothness for robot motion. Motion primitives based search space guarantees the smoothness of the robot motion. For time limited planning problems, ARA*\cite{Likhachev2003ARAAA} could give a suboptimal solution withhin suboptimality $\epsilon$ and if there is more time, the path can be optimize based on the previous suboptimal solution.
In this paper we are using a simple A* approach for path planning. Figure~\ref{ss} shows a search space build from motion primitives \cite{sbpl}.

There are two kinds of cost functions defined for the A* algorithm. $g(s)$ denotes the cost from the start point to the current point, the cost is the smallest path length from the start point to the current point. $h(s)$ denotes the heuristic function that estimates the cost from the current point to the target point. To guarantee A* to give the solution, $h(s) \le c(s, s^{\prime}) + h(s^{\prime})$ must be satisfied for any successor $s^{\prime}$, with  $c(s, s^{\prime})$ being the cost from the current point to the successor $s^{\prime}$.
\par

With such a MP A* planning approach similar states will be revisited quite often, coming from different paths. This is a disadvantage compared to grid-map based planning, resulting in long computation times. In order to speed up our planning, we define $\beta$ as a coefficient of penalty in a grid overlayed over the space. For example, if a path failed at the end, the cell covered by the path will be assigned a worse heuristic, since the potential of failure through this cell is high. The distance to the failure state decides how worse it gets. In other words, the closer to failure spot, the higher the probability to fail again through this state.

\begin{figure}[tb]
  \centering
  \begin{tikzpicture}[auto]
    \draw[step=1cm, gray, very thin] (-0.8, -0.8) grid (2.8,2.8);
    \pgftext[bottom, x=0.5cm, y=0.2cm] {$a$};
    \pgftext[bottom, x=1.5cm, y=0.2cm] {$b$};
    \pgftext[bottom, x=2.5cm, y=0.2cm] {$c$};
    \pgftext[bottom, x=0.3cm, y=0.6cm] {$P_{i}$};
    \node [obstacle, name=obs, rotate=90] at (3, 0.8) {};
    \draw[very thick, ->] (0.2,0.5) to (1.2,0.5);
    \draw[very thick, ->] (1.2,0.5) to (2.2,0.5);
    \draw[very thick, ->, red] (2.2,0.5) to (3.2,0.5);
    \draw[very thick, ->, red] (2.2,0.5) to [bend right] (3,0.8);
  \end{tikzpicture}
  \caption{A path $P_{i}$ through state $a$, $b$ and $c$ and collision will occur for $2$ end successors.}
  \label{penalty}
\end{figure}
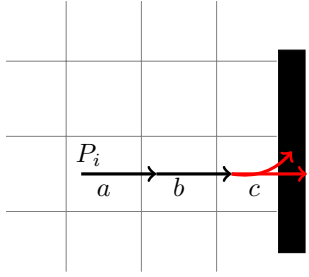

As shown in Figure~\ref{penalty}, a path $P_{i}$ from $a$ through $b$ ending with $2$ successors is colliding with an obstacle. In consequence, the state $c$ will get the largest punishment, because its direct successors failed. The state $b$ will get a slight punishment as well. Therefore, other successors will have less chance to be expanded. This heuristic design will encourage the different states from failure ones, leading to faster and more efficient planning. After pruning, obstacle-free paths have a high probability to be explored. Paths with high penalty indicate potential or obvious collision. Algorithm \ref{planning} shows this MP-based planning. The \textit{ExtractPathAction} is just back tracking the search process to find the optimal path, which is not illustrated in the paper. For checking visited cells and the termination condition checking, we apply cells into our implementation as in \cite{lavalle2006planning}.

\begin{algorithm}[tb]
  \caption{MP-Based Planning}
  \label{planning}
  \begin{algorithmic}[1] 
      \Require Motion Primitives $M$, Robot Initial and Goal Location $x_{0}$, $x_{G}$
      \Ensure Robot Path $P$, Robot Action $A$
      \Function {Plan}{$M$, $x_{0}$, $x_{G}$}
      \State openlist.empty(), closedlist.empty()
      \State $P=\emptyset $, $A=\emptyset$
      \State openlist.push($x_{0}$)
      \Do
        \State $x\in $ openlist and lowest cost
        \State Span motion tree using $M$
        \State closedlist.push($x$)
        \State openlist.pop($x$)
        \State openlist.push(all succesors of $x$)
      \doWhile {openlist  $\ne \emptyset$ or $x_{G} \in $ openlist}
      \State [$P$, $A$] = ExtractPathAction(closedlist)
      \State \Return{$P$, $A$}
      \EndFunction
  \end{algorithmic}
\end{algorithm}

\subsection{Adaptive Path Planning}

The planned path will be executed based on the steering commands saved in the motion primitives. We observe the result of the executed MPs to evaluate their performance. Each cluster of motion primitives will get a score from a voting system, which determines if the MP of that cluster will be used for path planning or not.

There could be three cases and there is a \textbf{C}onsistency \textbf{C}heck with \textbf{D}esigned \textbf{M}otion \textbf{P}rimitive (CCDMP) and a \textbf{C}onsistency \textbf{C}heck with \textbf{C}andidate \textbf{M}otion \textbf{P}rimitive (CCCMP):

\begin{itemize}
  \item Consistent with the designed motion primitive. Upvote the cluster.
  \item Inconsistent with the designed motion primitive and not appear in the candidate clusters. Downvote the cluster and  add a new, candidate cluster.
  \item Inconsistent with the designed motion primitive and appear in the candidate clusters. Downvote the cluster and upvote the candidate cluster.
\end{itemize}
\par

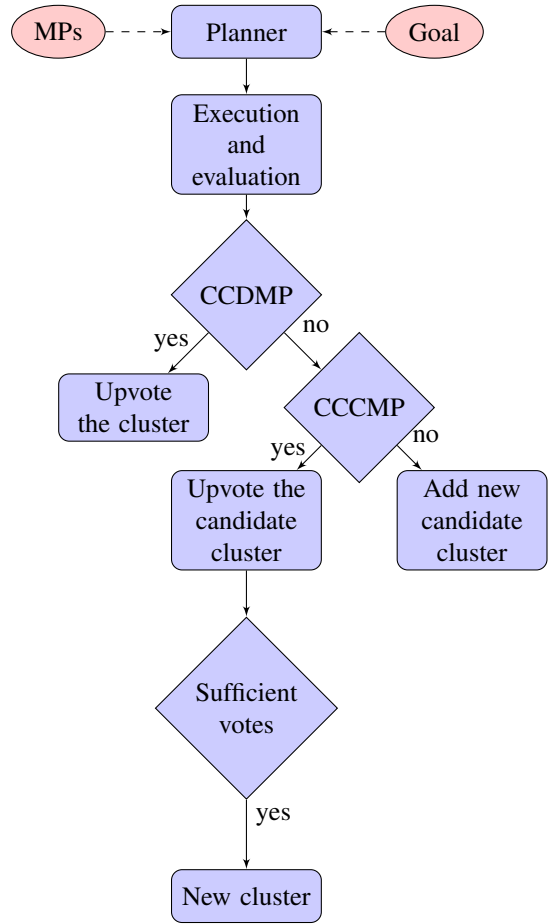
\begin{figure}[tb]
  \begin{tikzpicture}[node distance=1.5cm, auto]
    \node [block1] (plan) {Planner};
    \node [cloud, left of=plan] (mp) {MPs};
    \node [cloud, right of=plan] (goal) {Goal};
    \node [block1, below of=plan] (eva) {Execution and evaluation};
    \node [decision, below of=eva] (decide) {CCDMP};  
    \node [block1, below of=decide, left of=decide] (up) {Upvote the cluster};
    \node [decision, below of=decide, right of=decide, node distance=1.5cm] (down) {CCCMP};   
    \node [block1, below of=down, right of=down] (new) {Add new candidate cluster};
    \node [block1, below of=down, left of=down] (upcandidate) {Upvote the candidate cluster};
    \node [decision, below of=upcandidate, node distance=2.5cm] (evol) {Sufficient votes};
    \node [block1, below of=evol, node distance=2.5cm] (candidate) {New cluster};

    \path [line] (plan) -- (eva);
    \path [line] (eva) -- (decide);
    \path [line] (decide) -- node [near start, left] {yes} (up);
    \path [line] (decide) -- node [near start] {no} (down);
    \path [line] (down) -- node [near start, left] {yes} (upcandidate);
    \path [line] (down) -- node [near start] {no} (new);
    \path [line] (upcandidate) -- (evol);
    \path [line] (evol) -- node [near start] {yes} (candidate);
    \path [line,dashed] (mp) -- (plan);
    \path [line,dashed] (goal) -- (plan);
  \end{tikzpicture}
  \caption{The diagram of adaptive path planning}
  \label{diagram}
\end{figure}

Figure~\ref{diagram} shows the diagram of adaptive path planning. Every path planning, the motion primitives chosen from highest score of candidate clusters, including original generated motion primitives and new candidates after execution.

%

\section{Experiments}
\label{sec:experiments}
\begin{figure}[tb]
  \includegraphics[width=0.5\textwidth]{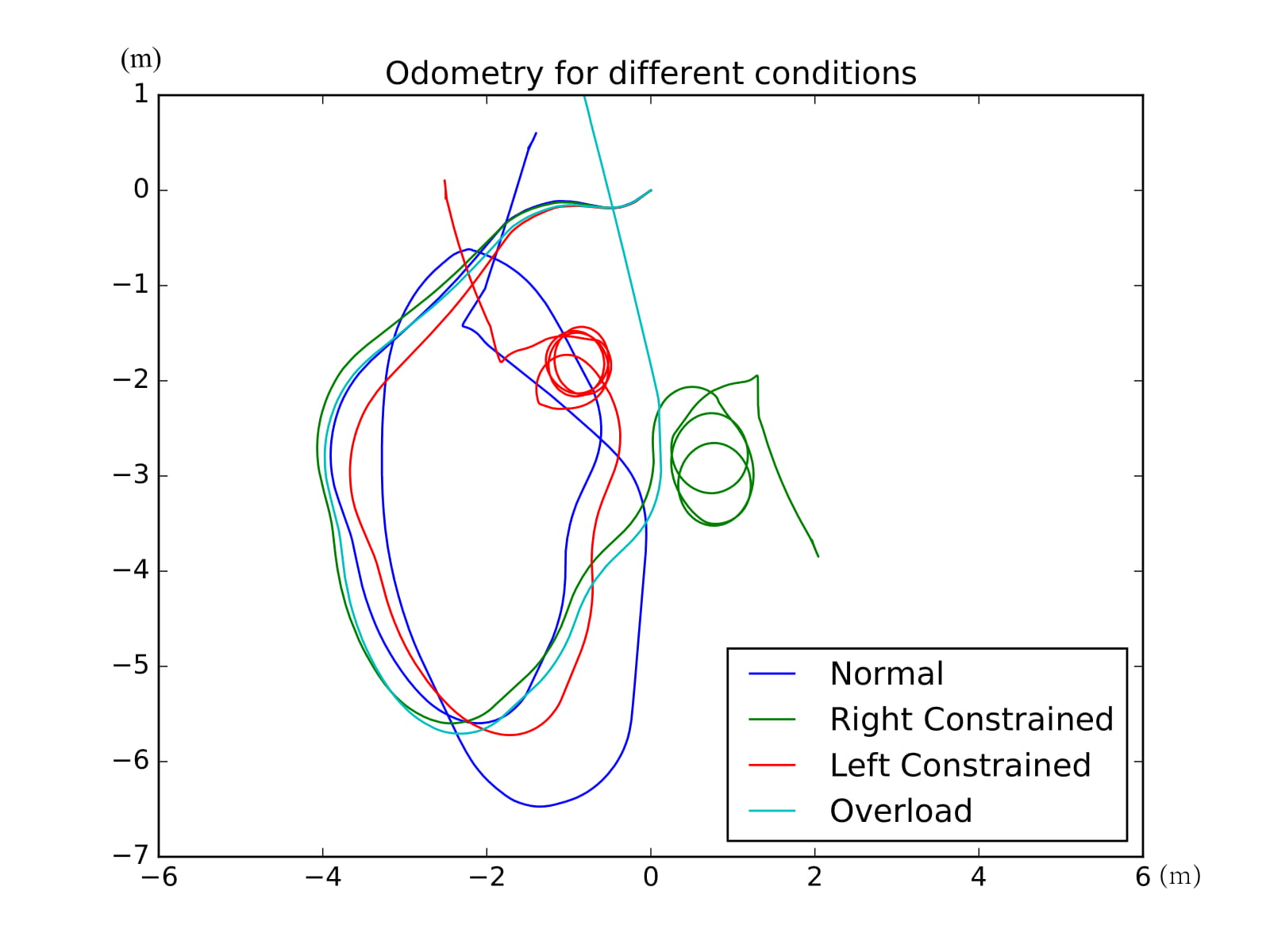}
  \caption{Odometry in different locomotion conditions.}
  \label{odoms}
\end{figure}

\begin{figure}[tb]
  \includegraphics[width=0.5\textwidth]{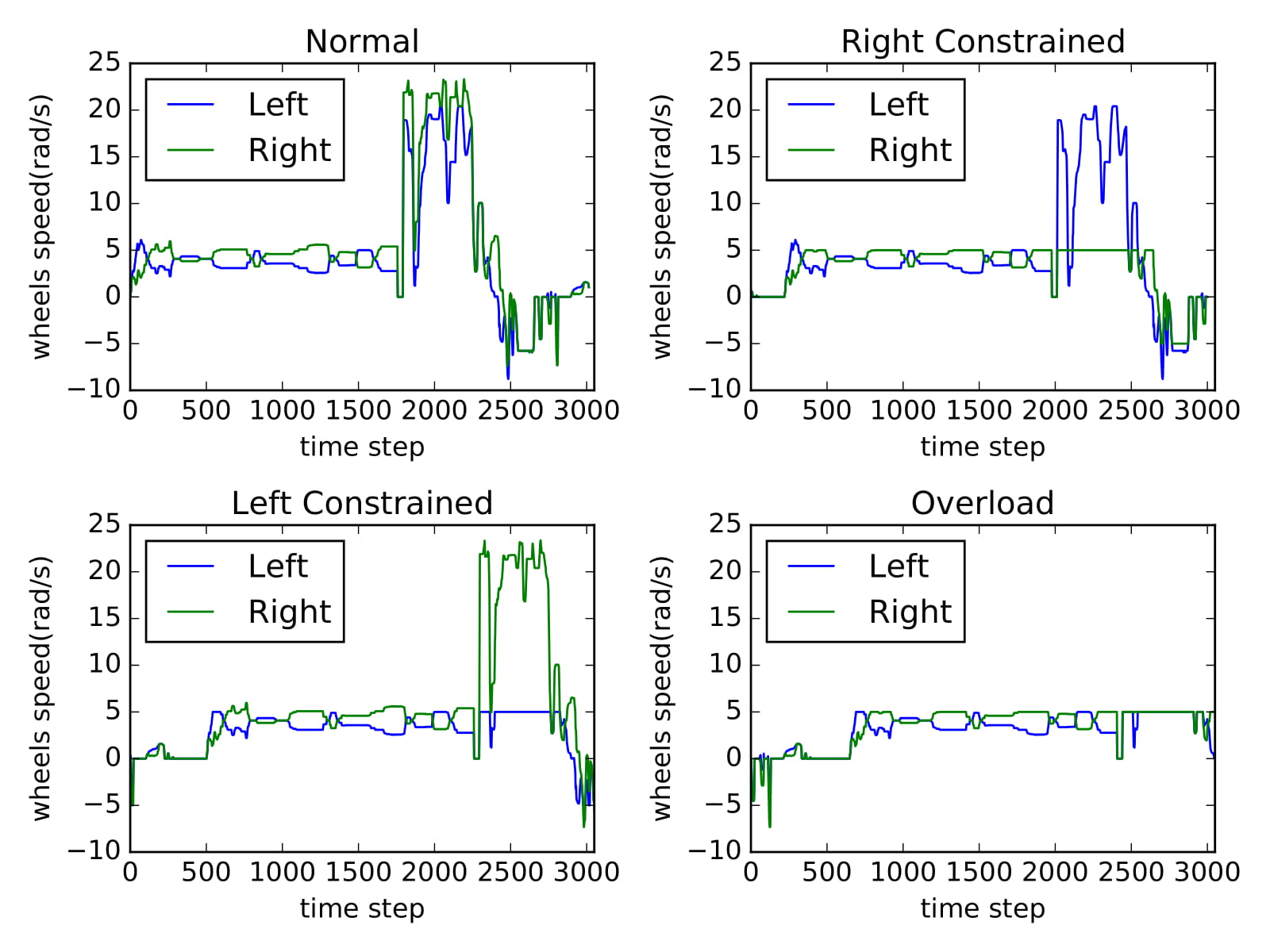}
  \caption{Wheel speed for different locomotion conditions.}
  \label{wheelspeed}
\end{figure}

In this paper, two experiments are used to evaluate the performance of the adaptive motion primitives based path planning. We take a look at balanced degraded locomotion conditions and at unbalanced locomotion conditions.
Firstly, we test the effect of three types of conditions as shown in Figure~\ref{odoms}. Unbalanced conditions means, that the locomotion on one side of the robot is degraded (e.g. loose axis, de-tracked) and thus the robot will move more into the degraded side. The unbalanced condition results in circular movement in place. Figure~\ref{wheelspeed} shows the wheel speeds of the four different cases.

\subsection{MP-Based Navigation} 
Here we illustrate the feasibility the motion primitives based navigation. The robot is the Clearpath indoor mobile robot Jackal, including a i5-4790s CPU, 8 GB memory, IMU, wheel encoder and a Velodyne VLP16 LIDAR sensor. In a 10m $\times$ 10m lab environment, we teleoperate the robot for 200 seconds. BLAM\footnote{\tt{https://github.com/erik-nelson/blam}} is used for localization, since BLAM does not need wheel odometry for localization and has a good performance in localization and mapping. There are totally 138 motion primitives generated. Each MP contains 5 samples from both control command input and result path output for the clustering feature of each motion primitive. After applying affinity propagation on the motion primitives with similarity, we obtain 40 motion primitives with much less similarity.

\subsection{Balanced Degraded Locomotion Conditions}
The balanced degraded locomotion conditions for differential mobile robot means, that the two sides of the robot suffer the same constraints. One typical situation is a too heavy payload. Once the robot (balanced) overloads, both left and right wheel speed are bound tighter. The robot is not able to move as fast as in the normal load condition.
Figure~\ref{overload} shows the planners expected wheel speed (green line) and the robots actual achievable wheel speed (blue line) under maximal wheels speed constrained to $5$ rad/s for both sides. The normal navigation is using move\_base\footnote{\url{http://wiki.ros.org/move_base}} with DWA as local planner. It is clear the wheel speed can not achieve the speed more than maximal wheel speed constrained. Figure~\ref{ol} shows the performance of normal navigation and our method, which the normal navigation move more in a line till it reaches the vertical goal. In contrast, our method works fine, except in the beginning of the motion. Since the normal navigation pursues to reach the goal quickly, it failed in this situation. The robot will go straight although it is supposed to turn left or right. Instead, our method chooses the best motion primitives with control input, evaluating the motion primitives during execution. After the algorithm detects any inconsistency in the motion primitives, it is able to adopt the new motion primitive, which is more suitable.

\newcommand\blaSize{1.9 in}

\newcommand\bluSize{1.5 in}

\begin{figure}[tb]
\centering

\subfigure[Normal Navigation to (3, 2)]{
\begin{minipage}[t]{0.2\textwidth}
\centering
\includegraphics[width=\bluSize]{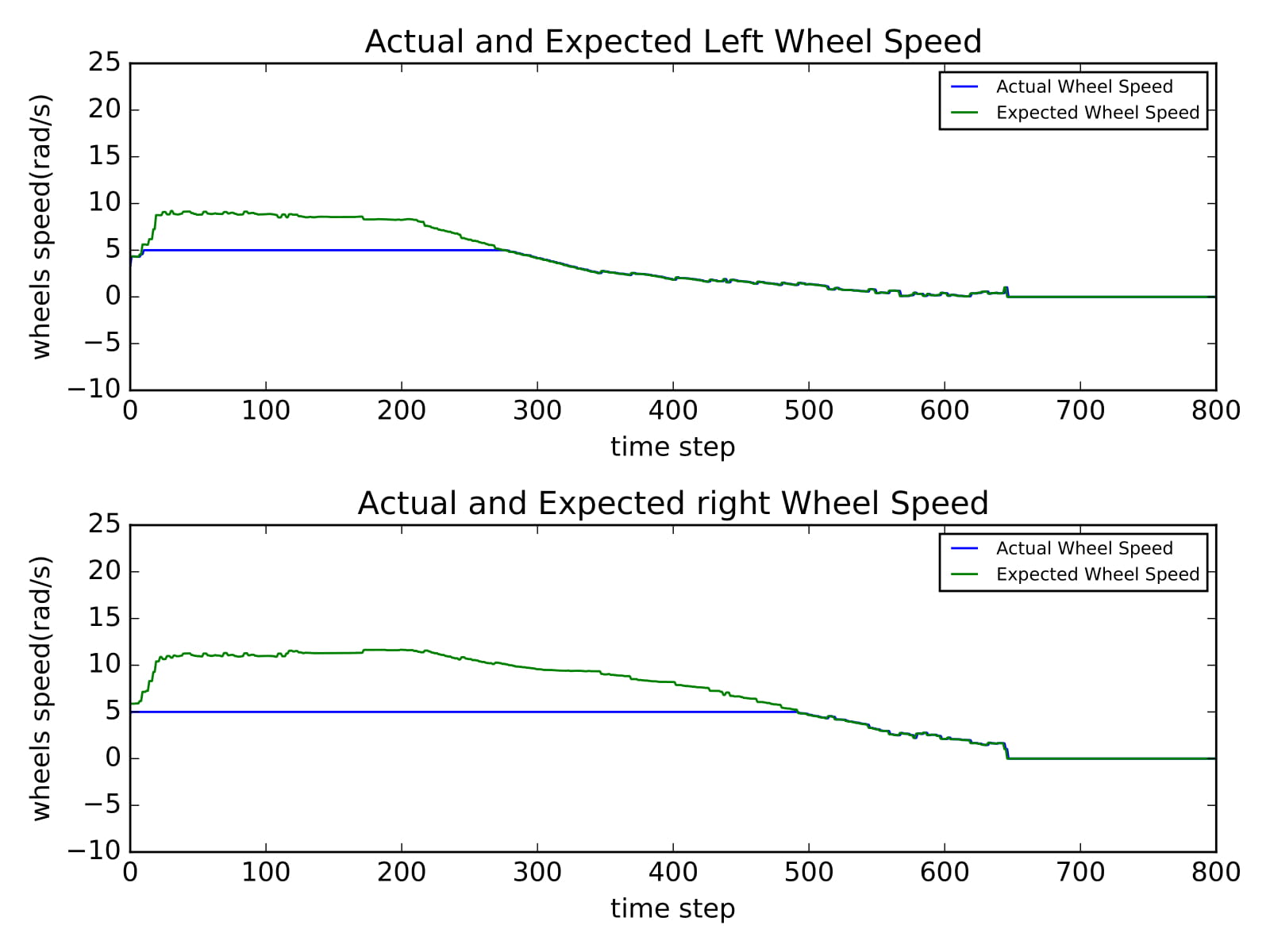}
\end{minipage}%
}%
\subfigure[Ours to (3, 2)]{
\begin{minipage}[t]{0.2\textwidth}
\centering
\includegraphics[width=\bluSize]{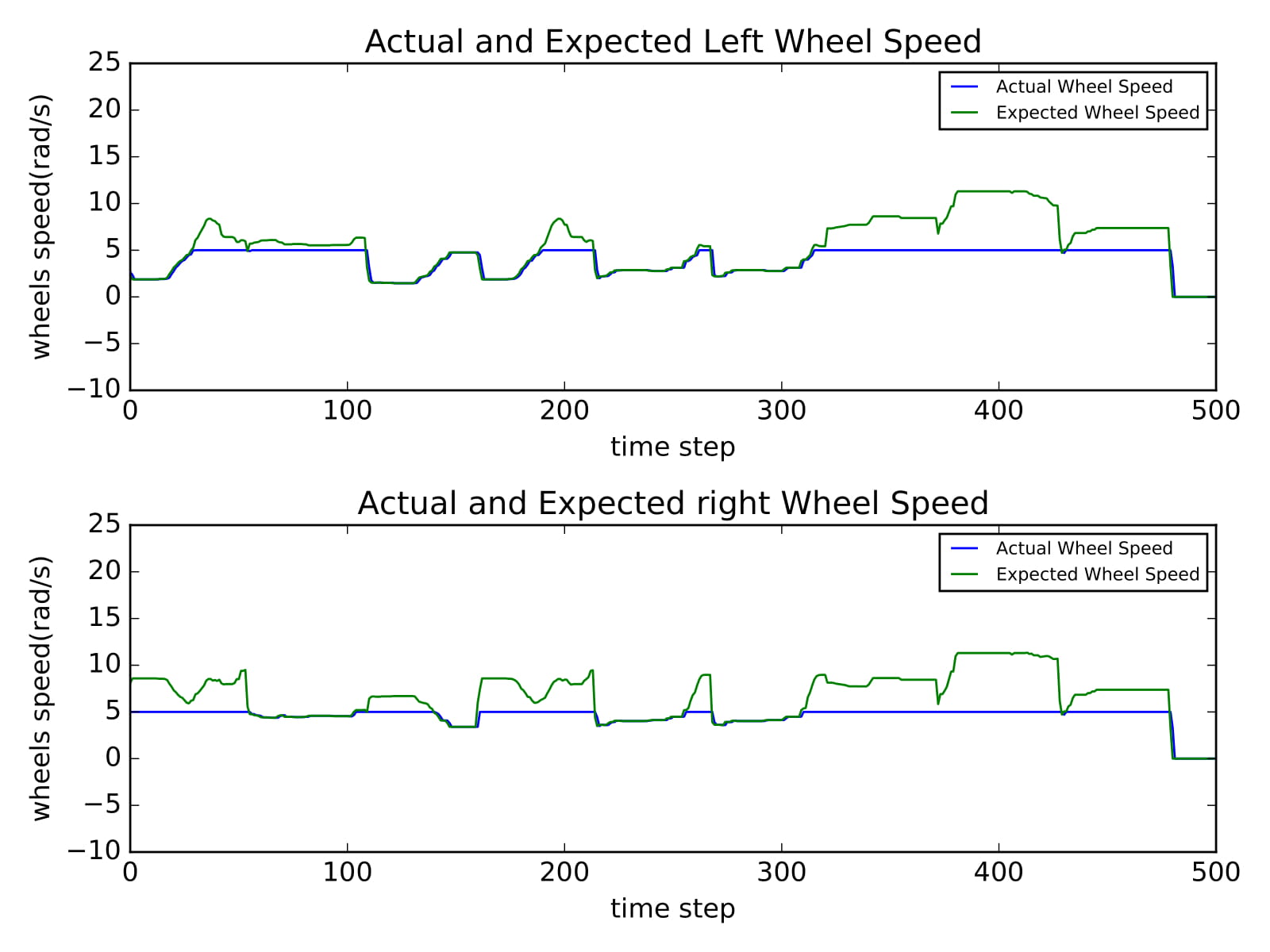}
\end{minipage}%
}%

\subfigure[Normal Navigation to (3, -2)]{
\begin{minipage}[t]{0.2\textwidth}
\centering
\includegraphics[width=\bluSize]{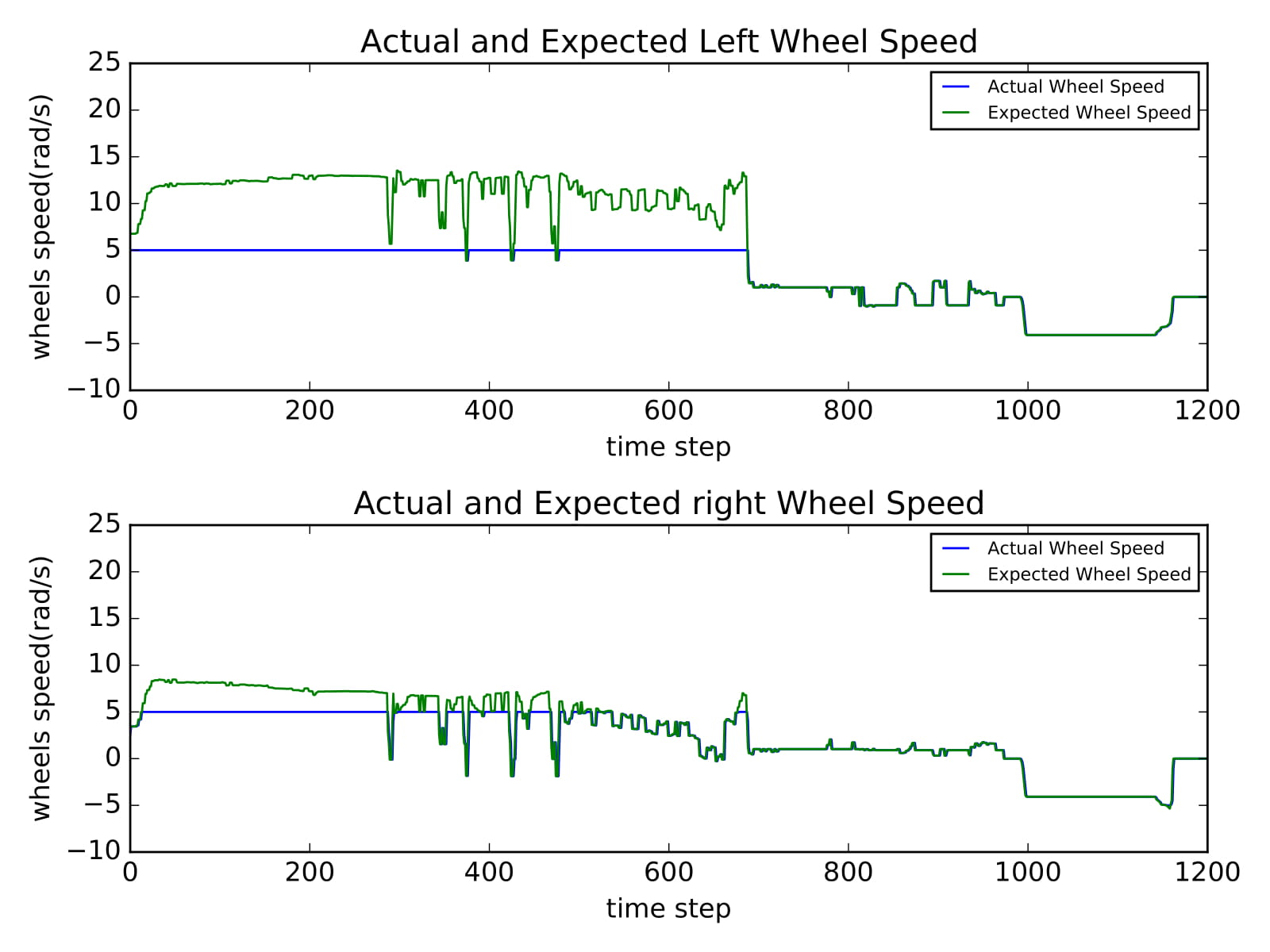}
\end{minipage}
}%
\subfigure[Ours to (3, -2)]{
\begin{minipage}[t]{0.2\textwidth}
\centering
\includegraphics[width=\bluSize ]{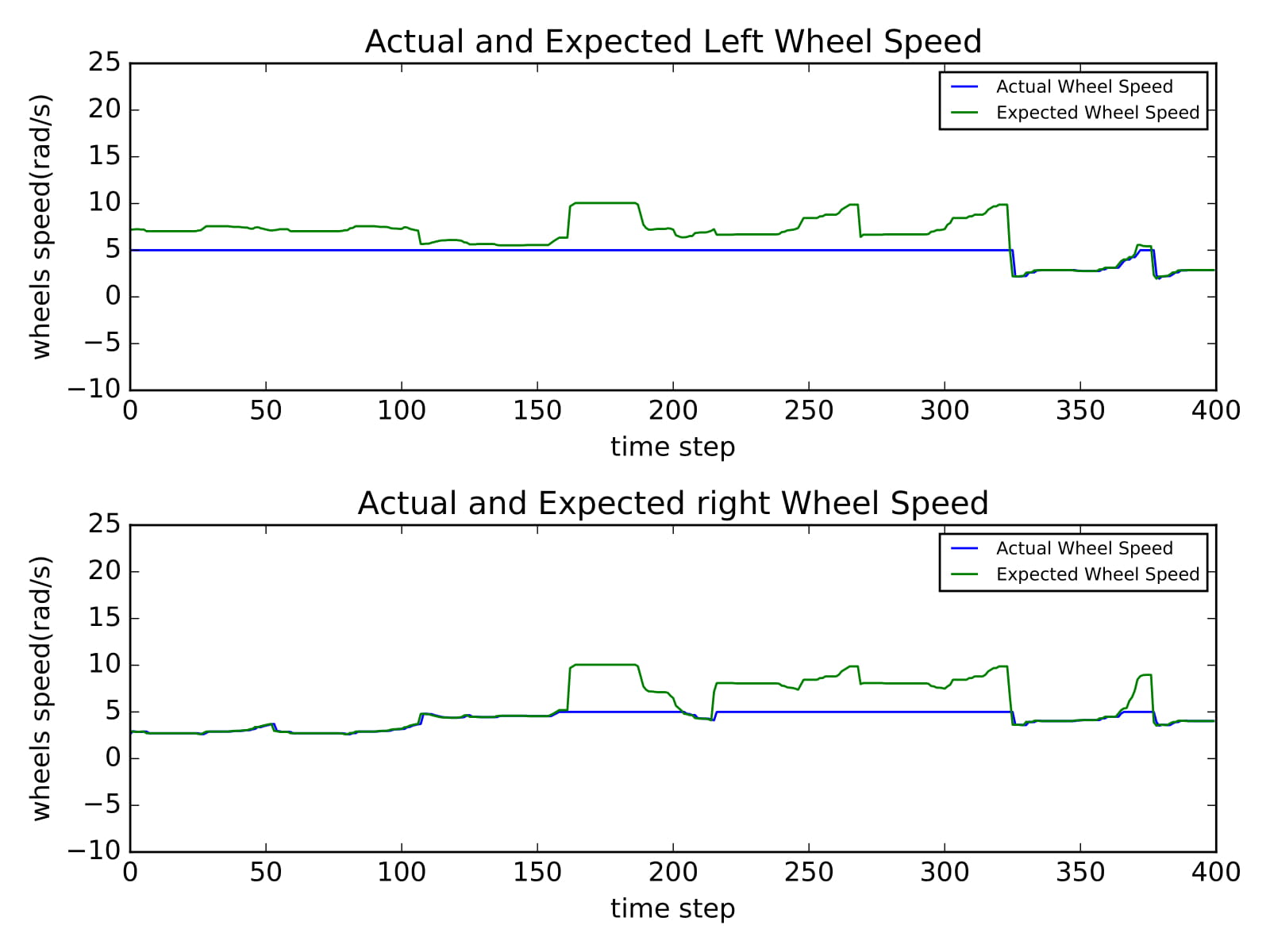}
\end{minipage}
}%

\centering
\caption{The expected wheel speed and the actual wheel speed for balanced degraded locomotion condition.}
\label{overload}
\end{figure}

\begin{figure}[tb]
\centering

\subfigure[Goal to (3, 2)]{
\begin{minipage}[t]{0.2\textwidth}
\centering
\includegraphics[width=\bluSize]{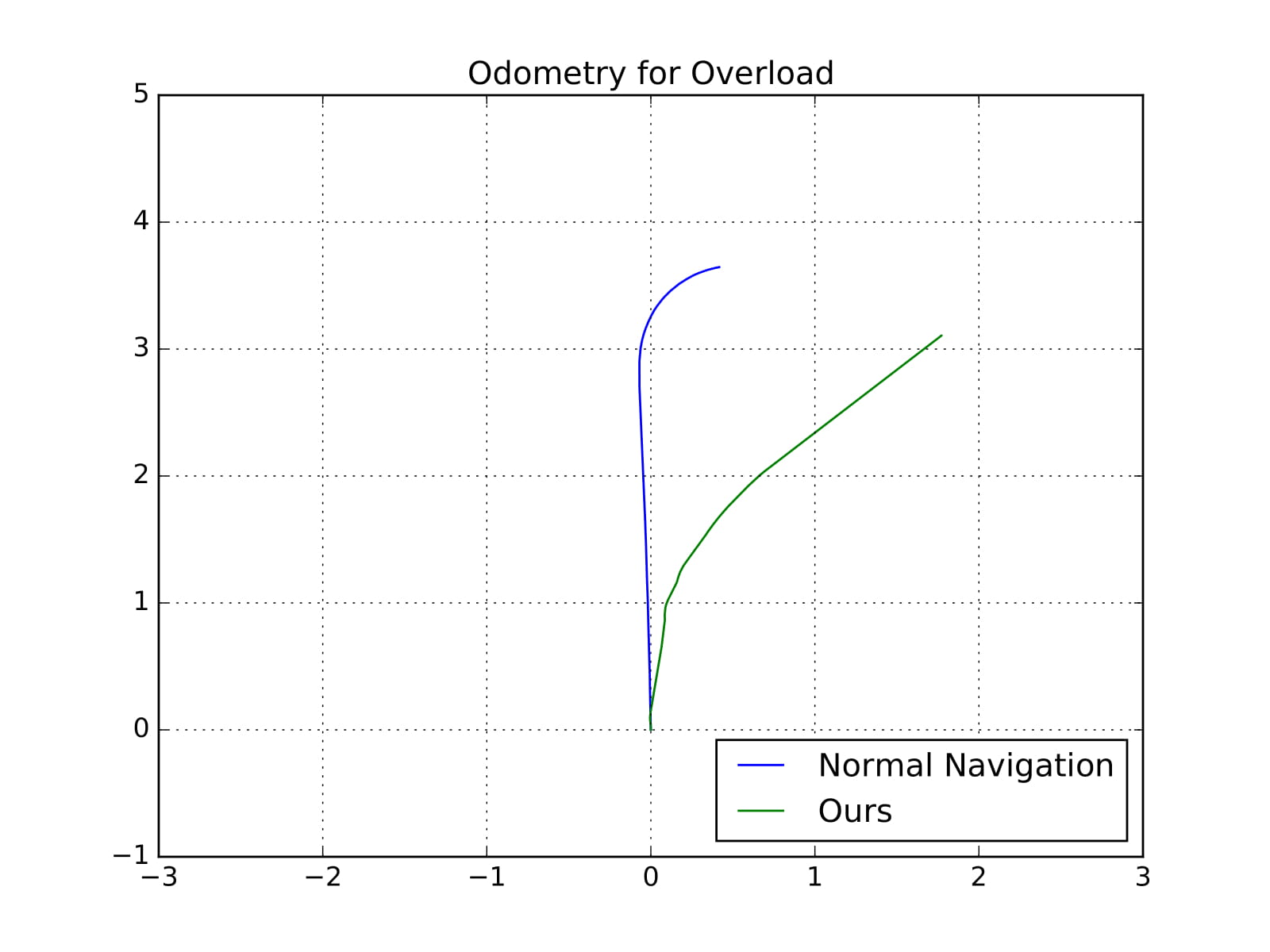}
\end{minipage}%
}%
\subfigure[Goal to (3, -2)]{
\begin{minipage}[t]{0.2\textwidth}
\centering
\includegraphics[width=\bluSize]{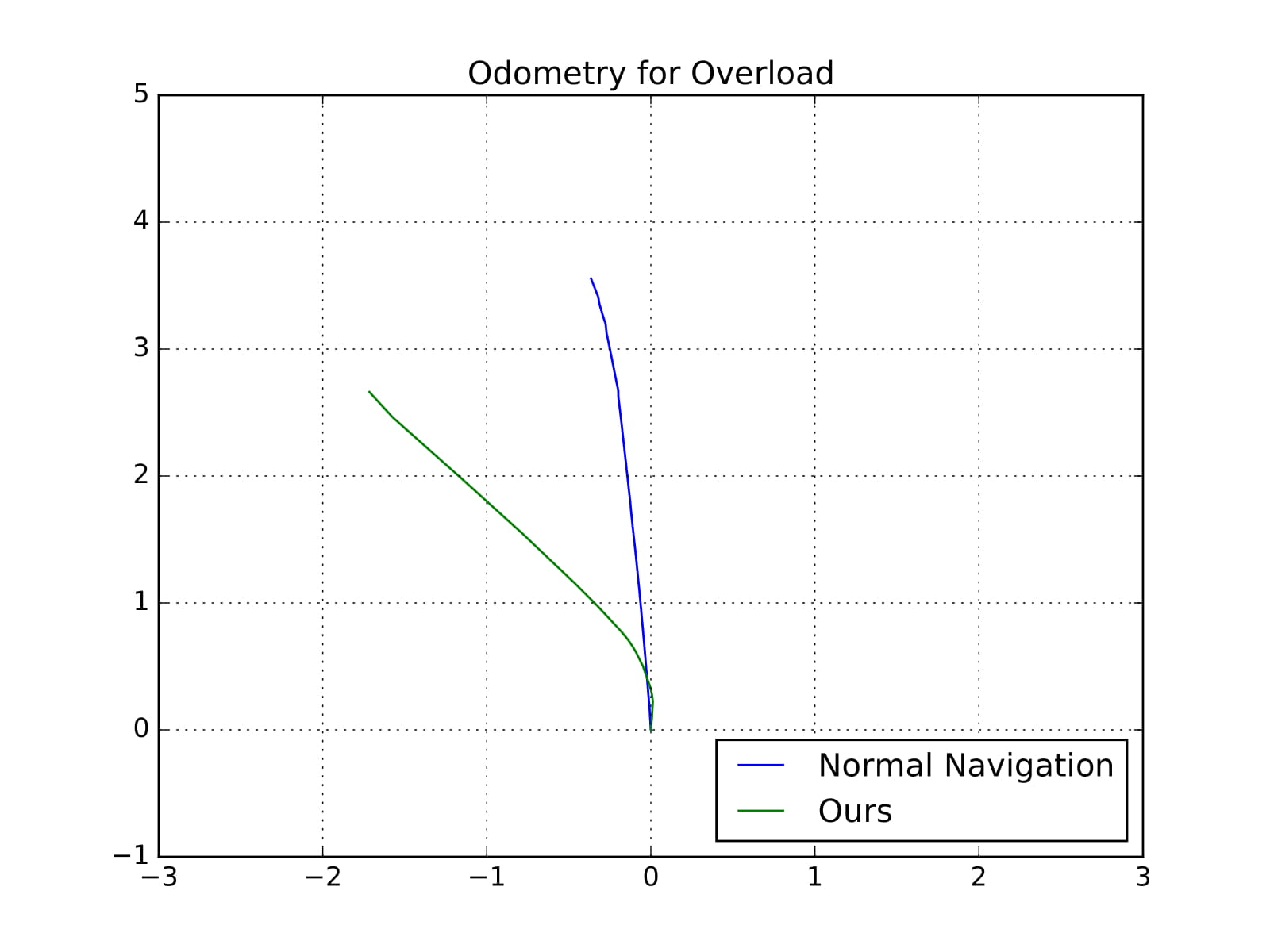}
\end{minipage}%
}%

\centering
\caption{The result path for normal navigation and our method for balanced degraded locomotion condition.}
\label{ol}
\end{figure}

\subsection{Unbalanced Degraded Locomotion Conditions}
\begin{figure}[tb]
\centering

\subfigure[Normal Navigation to (3, 2)]{
\begin{minipage}[t]{0.2\textwidth}
\centering
\includegraphics[width=\bluSize]{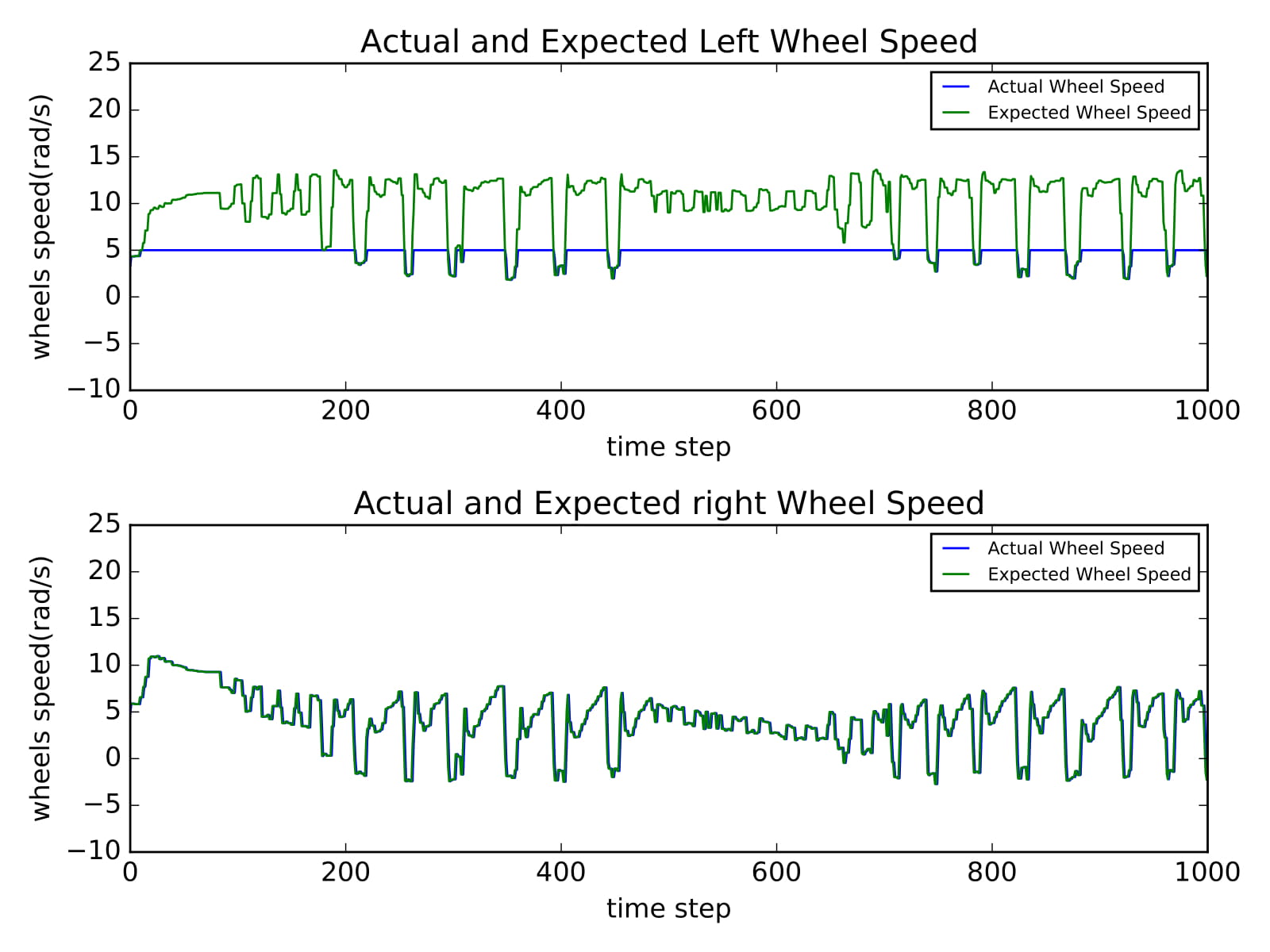}
\end{minipage}%
}%
\subfigure[Ours to (3, 2)]{
\begin{minipage}[t]{0.2\textwidth}
\centering
\includegraphics[width=\bluSize]{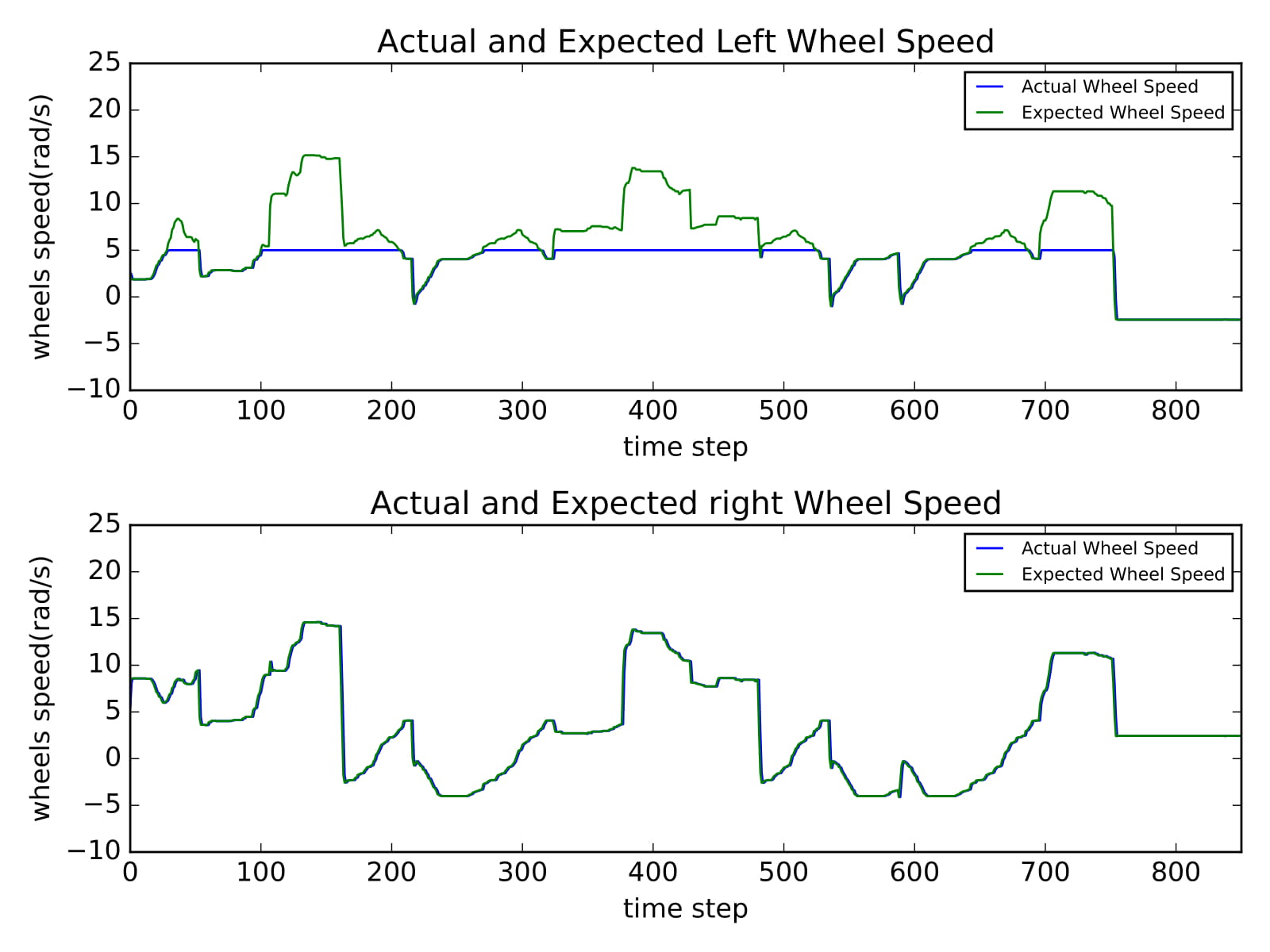}
\end{minipage}%
}%

\subfigure[Normal Navigation to (3, -2)]{
\begin{minipage}[t]{0.2\textwidth}
\centering
\includegraphics[width=\bluSize]{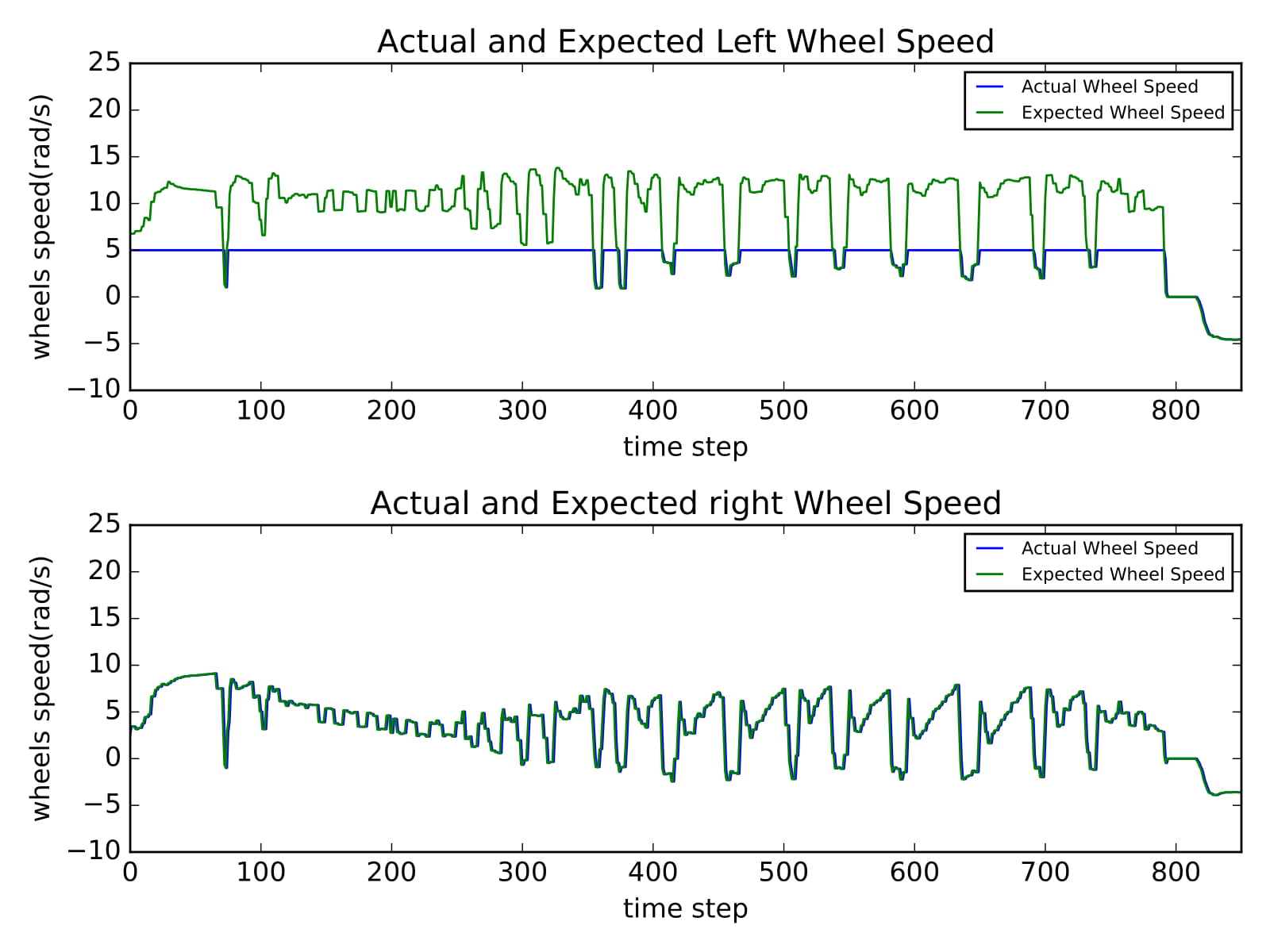}
\end{minipage}
}%
\subfigure[Ours to (3, -2)]{
\begin{minipage}[t]{0.2\textwidth}
\centering
\includegraphics[width=\bluSize]{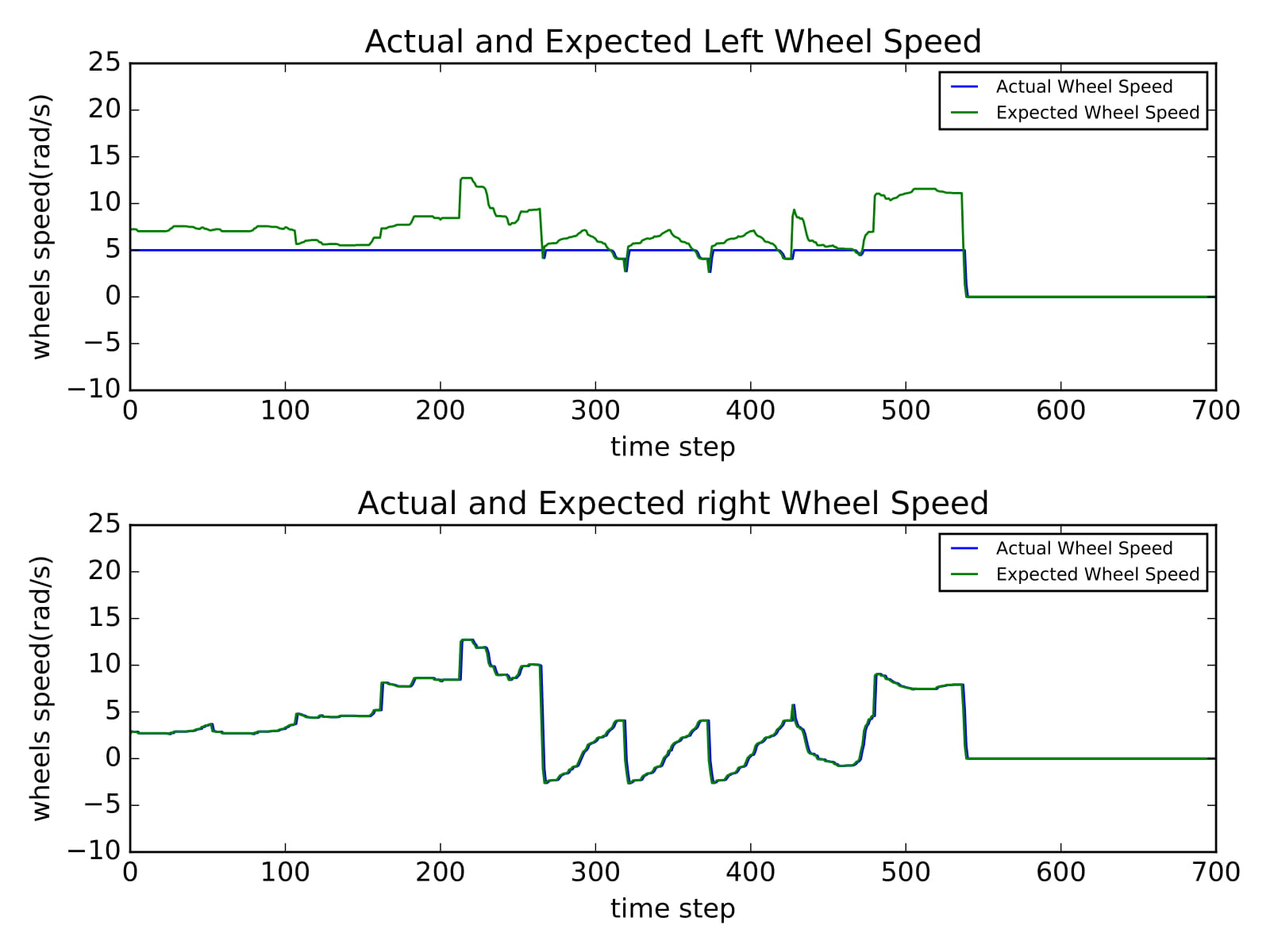}
\end{minipage}
}%

\centering
\caption{The expected wheel speed and the actual wheel speed for the unbalanced degraded locomotion condition.}
\label{left}
\end{figure}

\begin{figure}[tb]
\centering

\subfigure[Goal to (3, 2)]{
\begin{minipage}[t]{0.2\textwidth}
\centering
\includegraphics[width=\bluSize]{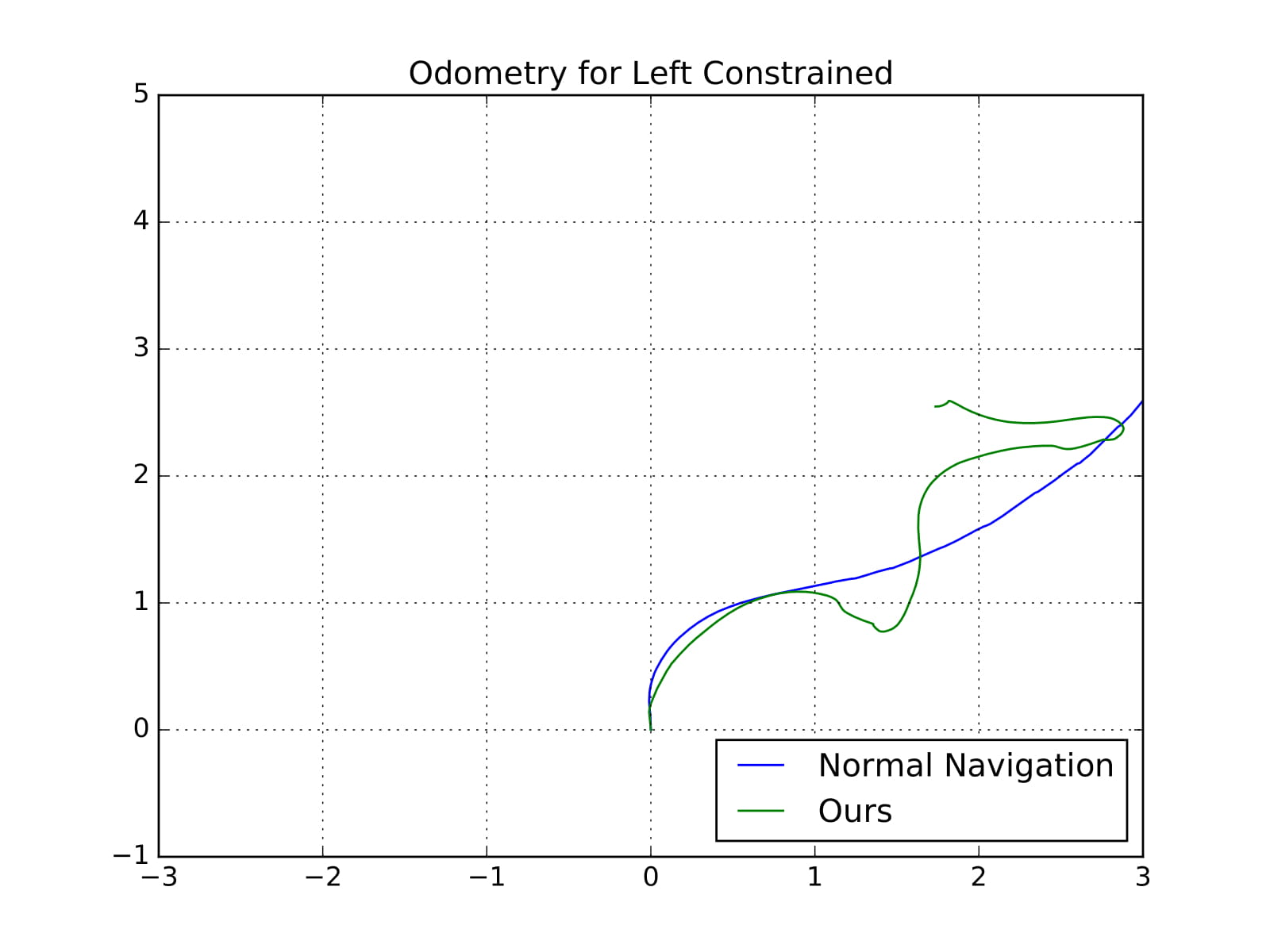}
\end{minipage}%
}%
\subfigure[Goal to (3, -2)]{
\begin{minipage}[t]{0.2\textwidth}
\centering
\includegraphics[width=\bluSize]{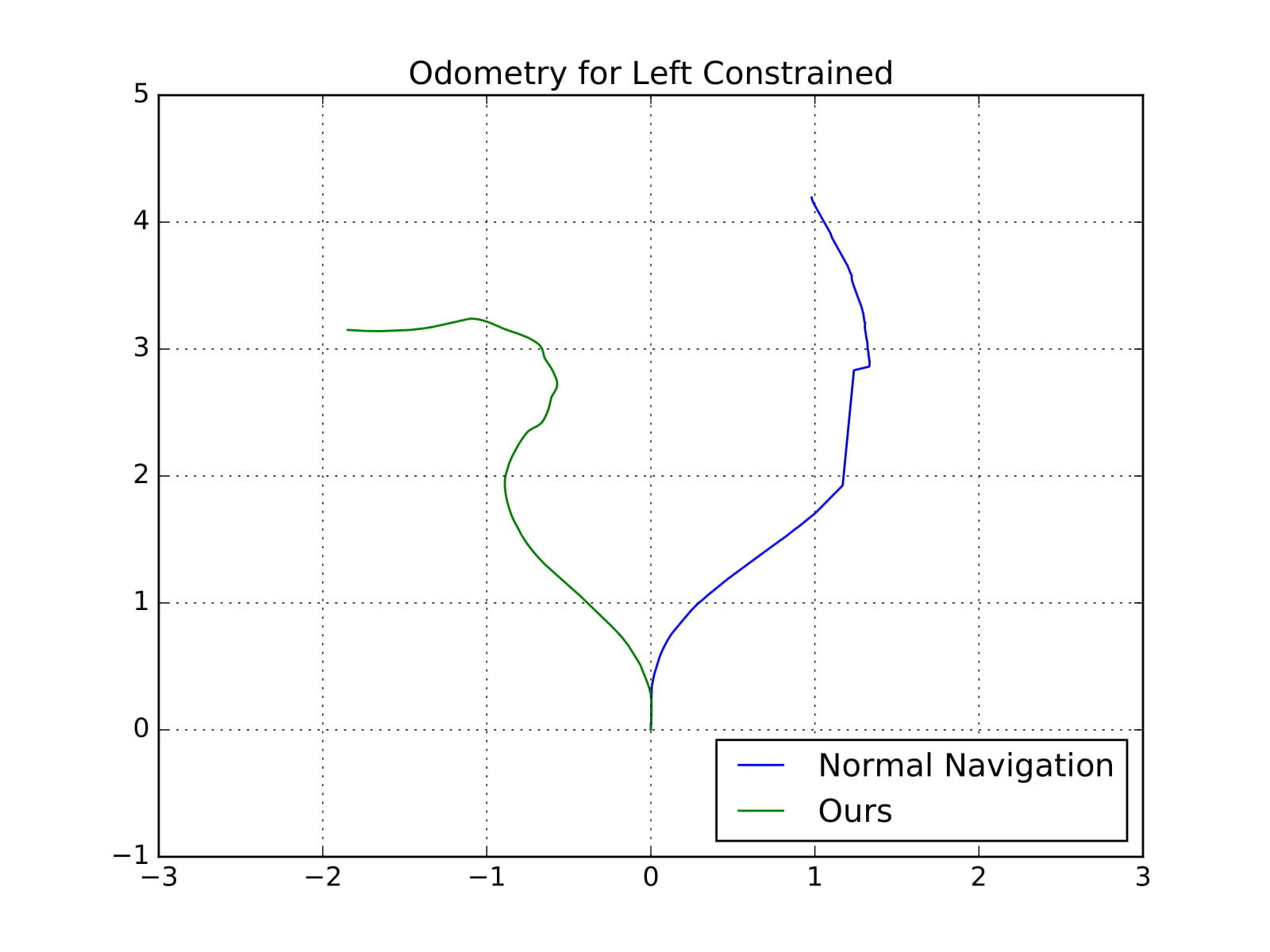}
\end{minipage}%
}%

\centering
\caption{The result path for normal navigation and our method for the unbalanced degraded locomotion condition.}
\label{leftodom}
\end{figure}

The unbalanced degraded locomotion conditions of robots have different causes, i.e. missing air in the wheel of one side or too much air in the other side.
Since the radius of the wheel changes, based on the kinematic model of the differential robot, the robot execution and wheel odometry have no good performance.
Figure~\ref{left} and Figure~\ref{leftodom} show the wheel speeds the planner expected (green line) together with the actual robot speed (blue line). For that we are using maximal wheel speeds constrained to $5$ rad/s for the left side and the result path from origin to goal $(3, 2)$ and $(3, -2)$ in both directions. The normal navigation uses the same configuration as mentioned above. Our method has not reached the goal point close enough, but if sufficient time for our method would be given, it could reach it closer, as it samples more MPs.

\section{Conclusions}
\begin{table*}[tb]
  \newcommand{\tabincell}[2]{\begin{tabular}{@{}#1@{}}#2\end{tabular}}
  \begin{center}
    \caption{Comparison between normal navigation and our method \label{table1}}
    \begin{threeparttable}
      \begin{tabular}{|c|c|c|c|c|c|c|c|}
      \hline
                                         & Methods                     & Goal    & Path Length(m) & End Point                & Run Time(s)    & Distance to Goal(m) & Result \\ \hline
      \multirow{4}{*}{Normal}            & \multirow{2}{*}{move\_base} & (3, 2)  & 3.35         & (3.05, 2.22)         & 6.87         & 0.22              & 1      \\ \cline{3-8}
                                         &                             & (3, -2) & 3.27         & (2.52, -1.81)        & 6.67         & 0.51              & 1      \\ \cline{2-8}
                                         & \multirow{2}{*}{ours}       & (3, 2)  & 3.91         & (3.17, 2.03)         & 7.62         & 0.27              & 1      \\ \cline{3-8}
                                         &                             & (3, -2) & 4.06         & (3.25, -1.78)        & 7.62         & 0.33              & 1      \\ \hline
      \multirow{4}{*}{Right Constrained} & \multirow{2}{*}{move\_base} & (3, 2)  & 6.01         & (5.25, -0.74)        & 15.06        & 3.54              & 0      \\ \cline{3-8}
                                         &                             & (3, -2) & 6.40         & (3.70, -3.27)        & 15.10        & 1.45              & 0      \\ \cline{2-8}
                                         & \multirow{2}{*}{ours}       & (3, 2)  & 5.20         & (3.65, 1.68)         & 15.01        & 0.73              & 1      \\ \cline{3-8}
                                         &                             & (3, -2) & 5.73  & (3.29, -2.71) & 15.01 & 0.77       & 1      \\ \hline
      \multirow{4}{*}{Left Constrained}  & \multirow{2}{*}{move\_base} & (3, 2)  & 7.12         & (5.10, 2.96)         & 17.05        & 2.31              & 0      \\ \cline{3-8}
                                         &                             & (3, -2) & 4.76         & (4.19, 0.98)         & 12.04        & 3.21              & 0      \\ \cline{2-8}
                                         & \multirow{2}{*}{ours}       & (3, 2)  & 6.04         & (2.55, 1.74)         & 17.00        & 0.52              & 1      \\ \cline{3-8}
                                         &                             & (3, -2) & 4.65         & (3.15, -1.84)        & 12.01        & 0.21              & 1      \\ \hline
      \multirow{4}{*}{Overload}          & \multirow{2}{*}{move\_base} & (3, 2)  & 3.86         & (3.65, 0.42)         & 10.04        & 1.71              & 0      \\ \cline{3-8}
                                         &                             & (3, -2) & 3.58         & (3.55, -0.36)        & 8.06         & 1.73              & 0      \\ \cline{2-8}
                                         & \multirow{2}{*}{ours}       & (3, 2)  & 3.74         & (3.11, 1.77)         & 10.05        & 0.25              & 1      \\ \cline{3-8}
                                         &                             & (3, -2) & 3.25         & (2.66, -1.72)        & 8.00         & 0.44              & 1      \\ \hline
      \end{tabular}
      \begin{tablenotes}
        \small
        \item In the Result column, a $1$ indicates Success and $0$ indicates Failure.
      \end{tablenotes}
    \end{threeparttable}
  \end{center}
\end{table*}

Table~\ref{table1} shows a comparison of the normal navigation and our method. When move\_base is trying to reach the goal under degraded conditions, it is failing. We choose the end time of our method as the end time to record data in the table. In the normal case, both move\_base and our method reach the goal and have similar performance. Under degraded conditions, move\_base can not reach the goal after lots of time. But our method does reach the goal. The result indicates that motion primitive based navigation takes the robot dynamic constraints into account. Instead of using forward kinematics to obtain these control-sample motion primitives, this paper generates motion primitives from the robot motion from the control input. We apply the affinity propagation clustering algorithm to reduce the similarity among these motion primitives. The result cluster center serves as the initial motion primitive to perform path planning.
In order to make path planning adaptive to different robot states and robot work environments,  the execution evaluates the behavior of the motion primitives. Based on the performance, we implement a voting system to weigh the quality of the motion primitives. Once the robot state or robot environment changes, the result motion primitives might be different than the initial ones, when comparing not only the result path, but also including the corresponding control command input. This evaluation maintains the candidate motion primitives for path planning, which adopts to degraded locomotion conditions.
\par
In the future, instead of pre-obtaining lots of motion primitives, the online clustering could have better result on most of degraded locomotion conditions. Online generation of motion primitives could be tolerant to any kind of movable locomotion conditions. Other navigation algorithms will be also used for further evaluation to reveal our algorithm's effectiveness.

\IEEEtriggeratref{13}

\bibliographystyle{IEEEconf}
\bibliography{root}

\end{document}